\documentclass[10pt]{article} 
\usepackage[preprint]{rlc}
\usepackage{graphicx} 
\usepackage{tikz}
\usepackage{amsmath}
\usepackage{nicematrix}
\usepackage{caption}
\usepackage{subcaption}
\usepackage{xspace}
\usepackage{natbib}
\usepackage{url}
\usepackage[normalem]{ulem}
\usepackage{booktabs}

\usepackage{amssymb}            
\usepackage{mathtools}          
\usepackage{mathrsfs}           
\mathtoolsset{showonlyrefs}     
\usepackage{graphicx}           
\usepackage{subcaption}         
\usepackage[space]{grffile}     
\usepackage{url}                
\newcommand{\jued}{\texttt{JaxUED}\xspace}
\newcommand{\minimax}{\texttt{minimax}\xspace}

\title{\jued: A simple and useable UED library in Jax}

\author{
    \centerline{Samuel Coward\thanks{Correspondence to \texttt{scoward@robots.ox.ac.uk}.} \qquad Michael Beukman \qquad Jakob Foerster} \\
    \centerline{FLAIR, Department of Engineering Science, University of Oxford}
}

\begin{document}

\maketitle

\begin{abstract}
    We present \jued, an open-source library providing minimal dependency implementations of modern Unsupervised Environment Design (UED) algorithms in Jax. \jued leverages hardware acceleration to obtain on the order of $100\times$ speedups compared to prior, CPU-based implementations.
    Inspired by CleanRL, we provide fast, clear, understandable, and easily modifiable implementations, with the aim of accelerating research into UED. 
    This paper describes our library and contains baseline results. Code can be found at \url{https://github.com/DramaCow/jaxued}.
\end{abstract}

\section{Introduction}

Reinforcement Learning (RL) in general has attracted much attention in recent years, leading to impressive results in several challenging domains~\citep{mnih2015Human,silver2017Mastering,vinyals2019AlphaStar}. 
More recently, Jax~\citep{jax2018Github} has become popular for RL, due to its ability to leverage hardware acceleration to speed up RL training by orders of magnitude~\citep{brax2021github,gymnax2022Github,lu2022Discovered,koyamada2023Pgx,sapora2023Evil}. 
Another subfield of RL that has seen growth is that of unsupervised environment design (UED), where an adversary generates environment configurations---known as levels---for a student to learn on~\citep{dennis2020Emergent,jiang2020Prioritized,jiang2021Replayguided,holder2022Evolving,mediratta2023Stabilizing}. By training on an adaptive curriculum, agents tend to learn faster and generalise better~\citep{jiang2021Replayguided,holder2022Evolving,team2023Humantimescale}.

Inspired by CleanRL~\citep{huang2021Cleanrl}---an RL library with understandable and single-file implementations of standard RL algorithms---we aim to accelerate research into UED by making high-quality implementations available and accessible to researchers. 
To this end, we introduce \jued: a fast, Jax-based UED library that contains (nearly) single-file implementations of common UED algorithms, all leveraging hardware acceleration to obtain significant speedups. Our implementations achieve on the order of $100\times$ speedup compared to prior CPU-based implementations, whilst maintaining evaluation performance comparable to existing implementations \citep{jiang2021Replayguided, jiang2023Minimax}. Additionally, unlike prior implementations, we find Domain Randomization (DR) performs competitively with current state-of-the-art UED methods on the common benchmark task of maze navigation.

\paragraph{Who is this for?}
\jued is primarily intended for researchers looking to get ``in the weeds'' of UED algorithm development. Our minimal dependency reference implementations expose the inner workings of the current state-of-the-art UED methods; helping researchers understand how the algorithms work in practice, and facilitating easy, rapid prototyping of new ideas. 
We are also inspired by another recent Jax-based UED library, \minimax~\citep{jiang2023Minimax}, which provides fast runtimes and strong baselines, in addition to multi-device training and reusable abstractions. \jued's focus is more on simple, single-file implementations of algorithms to facilitate rapid research into UED, and therefore prioritizes easily modifiable code over strict modularity and extensibility.

Our primary contributions are the following:
\begin{itemize}
    \item A design that prioritizes a minimal environment interface.
    \item Single-file reference implementations of common UED algorithms, allowing for quick and easy experimentation by researchers.
    \item Confirming the quality of our implementations by benchmarking against prior codebases.
    \item Contrary to past UED literature, we discover the surprising effectiveness of Domain Randomization.
\end{itemize}

\begin{table}[htbp]
\centering
\caption{Total wallclock time under each algorithm. The reported result is an average of 10 random seeds; however, the variance (for \jued) is negligible.
For DCD, we take the wall clock times from \citet{jiang2023Minimax}. We note the comparison is not entirely fair, since we used different hardware. However, we show it merely for comparison's sake, that \jued confers orders of magnitude speedups. For PAIRED, we take the number listed by \citet{jiang2023Minimax} and divide it by two since DCD performs double the total number of environment steps.}
\label{tab:wallclock_time}
\begin{tabular}{l|ccccc}
\hline
& \textbf{DR} & \textbf{PLR} & $\textbf{PLR}^\bot$ & \textbf{ACCEL} & \textbf{PAIRED} \\
\hline
dcd Wallclock Time (Hours) & 63 & - & 119 & 104 & 213 \\
\jued Wallclock Time (Hours) & 1.5 & 1.5 & 1.0 & 1.0 & 1.7 \\
\hline
\end{tabular}
\end{table}

\section{Unsupervised Environment Design}

Unsupervised Environment Design (UED) is a subfield of reinforcement learning concerning the unsupervised generation of sequences of environment distributions that facilitate the learning of robust policies \citep{dennis2020Emergent}. Formally, UED concerns Underspecified POMDPs \citep{dennis2020Emergent}, modelled as $\mathcal{M} = (A, O, \Theta, S, \mathcal{T}, \mathcal{I}, \mathcal{R}, \gamma)$, where: $A$ is the action space, $O$ is the observation space, $\Theta$ is the space of underspecified parameters referred to as \textit{levels}, $S$ is the set state space, $\mathcal{T}: S \times A \times \Theta \rightarrow \Delta(S)$ is the level-conditioned transition function, $\mathcal{I}: S \rightarrow O$ is the state to observation mapping, $\mathcal{R}: S \rightarrow \Delta(\mathbb{R})$ is the reward function, $\gamma$ is the discount factor.

UED is framed as a two-player game where a student policy is tasked with maximizing the discounted return on levels generated by an adversarial level generator. The adversary is tasked with generating levels that maximize some objective. Under this framework, Domain Randomization (DR) \citep{tobin2017Domain} can be viewed as a UED method whereby the adversary's utility for each level is constant; as such, the adversary merely has to present the student with levels sampled uniformly from $\Theta$. \citet{dennis2020Emergent} propose that the adversary generate levels that maximize the student's regret; that is, the difference between the expected discounted return achieved by an optimal policy on some level and the expected discounted return achieved by the student on the same level. Broadly, this has led to two classes of methods: 
\begin{itemize}
    \item PAIRED-based methods~\citep{dennis2020Emergent}: in which the adversary is itself an RL policy that generates levels, and is optimized to maximize the regret estimated by the difference between two (or more) student policies.
    \item Replay-based methods~\citep{jiang2020Prioritized, jiang2021Replayguided, holder2022Evolving}: whereby the adversary is represented by a rolling buffer of previously encountered levels with high regret estimates. These levels are discovered either through random search or evolution and the rolling buffer is periodically updated using the most recent regret estimates achieved on replayed levels.
\end{itemize}
\texttt{JaxUED} provides utilities for implementing these two classes of methods, and we provide concrete implementations for PAIRED, PLR, PLR$^\top$, ACCEL, in addition to DR.

\section{The \textsc{\jued} Library}

\textsc{\jued}'s design takes heavy inspiration from CleanRL~\citep{huang2021Cleanrl};
consequently, the amount of library code is minimal. In this section, we describe our core library features and in the next section, we discuss our reference implementations.

\subsection{Environment Interface}

RL libraries commonly implement environment interfaces that reflect Partially-Observable Markov decision processes. Without loss of generality, this includes (a) a \texttt{step} function that models some stochastic transition function; and (b) a \texttt{reset} function that models the initial state distribution. However, UED operates over Underspecified POMDPs~\citep[UPOMDP]{dennis2020Emergent}, which can be viewed as a collection of POMDPs whereby a specific POMDP is instantiated by some set of free parameters (aka. a level) $\theta$. Crucially, UPOMDPs do not define a ground truth distribution over levels; in fact, the role of UED is to adapt the level distribution over the course of training. This implies that prior environment interfaces are unsuitable for UED; if a developer were to implement a UPOMDP using prior interfaces, they would have to implicitly impose a distribution over levels.

This motivates the need for a new environment interface that more closely models UPOMDPs. As such, we introduce the \texttt{UnderspecifiedEnv}, a minimal environment interface that replaces the idea of a \texttt{reset} function, which would otherwise encode an implicit level distribution, with an explicit \texttt{reset-to-level} function. Consequently, \texttt{UnderspecifiedEnv} decouples the notion of level distribution from environments, offloading the management of level distributions to the external user, e.g., some UED algorithm, evaluation routine, etc. 

Additionally, we explicitly decouple the notion of levels from states: levels act as a context that induces a distribution over the state space. This is a strictly more general notion, as this distribution could be a Dirac delta function, which recovers the one-to-one correspondence between levels and states.

Our \texttt{UnderspecifiedEnv} interface defines the following methods:
\begin{itemize}
    \item \texttt{step}: This takes in an environment state and action (given by an external agent), and stochastically transitions to the next state before yielding an observation, reward, and termination flag.
    \item \texttt{reset\_to\_level} - which takes in a level $\theta$ and stochastically inititializes the environment state, returning an initial observation.
\end{itemize}

\subsection{Wrappers}

Note, that by decoupling the notion of level distribution from the environment interface, we cannot support automatic resetting of the environment state upon episode termination by default. In practice, however, automatically resetting is desirable for training. As such, we support automatic resetting through environment wrappers which transform a \texttt{UnderspecifiedEnv} to another \texttt{UnderspecifiedEnv}, inheriting behaviour where appropriate. These include:
\begin{itemize}
    \item \texttt{AutoReplayWrapper}: Upon episode completion, it will reset the environment state to some state sampled from the initial state distribution induced by the previously played level.
    \item \texttt{AutoResetWrapper}: Upon episode completion, this wrapper will first sample a new level from some predefined level distribution, then reset the environment state to some state sampled from the initial state distribution induced by the sampled level. 
\end{itemize}
These wrappers enable users to explicitly select automatic resetting behaviour for their particular use cases via dependency injection.

\subsection{Level Sampler}

Several UED methods implement a dynamic level distribution via a curated level buffer~\citep{jiang2021Replayguided,jiang2020Prioritized,holder2022Evolving}. As such, we provide an implementation of a \texttt{LevelSampler}, a rolling buffer of levels that associates each level with a score (i.e., regret estimate) and staleness (time since the level was last inserted or sampled). \texttt{LevelSampler} supports:
\begin{itemize}
    \item Sampling replay decisions, i.e., whether new levels should be evaluated or previous levels should be trained on.
    \item Inserting (a batch) of levels with associated scores.
    \item Updating (a batch) of levels with associated scores.
    \item Optional de-duplication, whereby attempted insertion of levels into the level sampler will instead update the score of the already existing level
    \item Sampling a batch of levels according to a distribution induced by level scores and staleness~\citep{jiang2020Prioritized}.
\end{itemize}
At times, users may wish to associate each level in the level buffer with auxiliary data. For example, each level may be associated with the largest return achieved during training (useful for certain regret estimates). To support this, each level is associated with an arbitrary dictionary called \texttt{level\_extra}. Such a feature is invaluable to those wishing the extend replay-based methods.

\section{Maze Environment}
We further provide a maze environment to showcase our reference implementations, as mazes are common benchmarks for UED~\citep{dennis2020Emergent,jiang2021Replayguided,holder2022Evolving,jiang2023Minimax}. We specifically provide:
\begin{itemize}
    \item \textbf{Maze environment}: A fully-JAX implementation of a simplified Minigrid environment, compliant with the \texttt{UnderspecifiedEnv} interface, whereby a partially observable agent is tasked with navigating to a goal position. Levels correspond to wall configurations as well as goal and agent start positions. Much like prior UED work~\citep{dennis2020Emergent, jiang2020Prioritized}, this environment yields observations consistent with the original MiniGrid implementation~\citep{MinigridMiniworld23}.
    \item \textbf{Efficient rendering}: Fully JIT-compiled image rendering capabilities for efficient visualization of generated levels, and agent animations.
    \item \textbf{JIT-compiled shortest-path}: a simple and easily extendable algorithm for (pre-)computing the shortest path to the goal from all agent positions, running in $O(N^2)$ for a level containing $N$ grid cells.
    \item \textbf{Maze Editor environment}: for UED methods that utilize an RL level editor policy (e.g., PAIRED), we supply a fully-JAX implementation of a Maze editor environment, compliant with the \texttt{UnderspecifiedEnv} interface. This environment is tasked with sequentially constructing a (potentially initially empty) level via atomic modifications (i.e. moving the agent or goal or adding and removing walls).
    \item \textbf{Level generation \& mutations}: Fully JIT-compiled level generation (for DR and PLR-based methods) and parameterized level mutation (for ACCEL) callbacks.
\end{itemize}

\section{Reference Implementations}
We now discuss our reference implementations, Domain Randomisation~\citep[DR]{jakobi1997Evolutionary,tobin2017Domain} , Prioritized Level Replay~\citep[PLR]{jiang2020Prioritized,jiang2021Replayguided}, ACCEL~\citep{holder2022Evolving} and PAIRED~\citep{dennis2020Emergent}.

\subsection{Replay-Based Methods}
We have one file that implements PLR, Robust PLR and ACCEL. This file has three primary subroutines, \texttt{on\_new\_levels}, \texttt{on\_replay\_levels} and \texttt{on\_mutate\_levels}; each encoding the different kinds of PPO updates (referred to as update-cycles) performed during replay-based UED methods.

\texttt{on\_new\_levels} generates a set of random levels and rolls the agent out on these levels for a fixed number of steps. These trajectories are then used to compute the score for each level (such as Positive Value Loss (PVL) or Maximum Monte Carlo (MaxMC)), and levels with high scores are added to the level buffer. PLR also updates the agent's policy on the trajectories from random levels, whereas robust PLR does not~\citep{jiang2021Replayguided}.

\texttt{on\_replay\_levels} only occurs when the level buffer is filled past a certain threshold (50\% by default). It samples a set of levels from the buffer, according to their regret scores and staleness. The agent then trains on these levels by first rolling out on them for a fixed number of steps, and thereafter updating the agent's policy.

\texttt{on\_mutate\_levels} only is chosen when ACCEL mode is activated \textit{and} the previous step was \texttt{on\_replay\_levels}. This function selects the previous batch of replayed levels, and randomly mutates them. The agent is rolled out on these to compute the regret scores for these children levels, and they are added to the buffer if their scores are sufficiently high.

Training thus simply consists of iteratively performing a fixed number of update cycles; we refer the reader to Figure~\ref{fig:replay} for more detail on which kind of update-cycles are performed each iteration.

\begin{figure*}[h]
    \centering
    \begin{subfigure}{0.45\linewidth}
    \begin{tikzpicture}[
        ->,>=stealth,shorten >=1pt,auto,node distance=3cm,
        state/.style={circle,draw,minimum size=1.5cm}
    ]
    
        \node[state] (A) {$A$};
        \node[state] (B) [right of=A] {$B$};

        \draw[->] (-2cm,0) -- (A);
        
        \path (A) edge [loop above, ->] node {\texttt{DR}} (A)
              (A) edge [bend left, ->] node[midway, above] {\texttt{Replay}} (B)
              (B) edge [->] node[midway, below] {\texttt{DR}} (A)
              (B) edge [bend left=40, ->] node[midway, below] {\texttt{Replay}} (A)
              (B) edge [bend left=90, ->] node[midway, below] {\texttt{Mutation}} (A);
    \end{tikzpicture}
    \caption{The state machine describing the canonical operations of replay-based methods. Training consists of two states, where the operation performed on the next update cycle is determined by a fixed, stochastic meta-policy.}
    \end{subfigure}
    \hfill
    \begin{subfigure}{0.45\linewidth}
        \centering
        \begin{align}
            \begin{bNiceMatrix}[first-row,first-col]
            &\texttt{DR} & \texttt{Replay} & \texttt{Mutation} \cr
            A & 1-p & p & 0 \cr
            B & (1-p)(1-q) & p(1-q) & q
            \end{bNiceMatrix}
    \end{align}
        \caption{The canonical replay-based meta-policy that stochastically selects what update-cycle to perform next based on the current training stage. $p$ and $q$ are hyperparameters corresponding to the \textit{replay probability} and \textit{mutation probability} respectively. Note, when using ACCEL, it is typical to select $q = 1$, as such a mutation update-cycle is always performed immediately after a replay update-cycle. When not using ACCEL, $q = 0$.}
        \label{fig:enter-label}
    \end{subfigure}
    \caption{Illustration of the training process of replay-based methods. Training can be viewed as a deterministic Markov Decision Process (MDP) controlled by a fixed, stochastic meta-policy defined by hyperparameters. Transitions determine what kind of PPO update is performed.}
    \label{fig:replay}
\end{figure*}

\subsection{Domain Randomisation}
Our domain randomisation implementation is similar to the PureJaxRL~\citep{lu2022Discovered} style of training, where multiple rollouts of the same policy are performed on different levels; these trajectories are used to update the policy.
Furthermore, by contrast with prior implementations~\citep{jiang2023Minimax}, we decouple our DR code and our PLR code. The reason for this is how trailing episodes are handled. In PLR-based methods, a certain number of levels are sampled, and a certain number of steps are performed on each. In our implementation, if the episode ends, the same level is used to reset the environment. This allows us to potentially have multiple episodes in each level, improving the regret score estimate. In DR, by contrast, each time an episode terminates, we wish to reset to a new level. If we only run a fixed number of timesteps, and reset to a new batch of levels at every iteration (as is done in PLR), we would lose the data from trailing episodes that have not been completed in the previous iteration. Since this is not how standard RL (which is effectively DR) is done, we follow the standard approach in our DR implementation, instead of using the (potentially flawed) approach of PLR.
\subsection{PAIRED}
PAIRED~\citep{dennis2020Emergent} has three agents, an adversary, an antagonist and a protagonist. The latter two agents are collectively referred to as students. At every iteration, we roll the adversary out in a level editor environment to generate a new set of levels. We then run both students on the same levels, and compute their rewards. The regret for each level is then estimated as the maximum antagonist performance minus the mean protagonist performance. The adversary is updated using a sparse reward of regret, and the students are updated using the reward as normal.\footnote{\citet{dennis2020Emergent} also experimented with using regret as the students' objective, but we choose reward.}

\section{Results}

In this section, we benchmark our implementations. In efforts to conduct a fair comparison between methods, we consider performance and wall-clock time of each method with respect to a fixed budget of environment interactions---instead of keeping the number of PPO updates constant as is sometimes done~\citep{holder2022Evolving}. For each methods, the number of environment steps is given as:
\begin{itemize}
    \item \textbf{DR}: $T$-step rollout length $\times$ number of parallel agents $\times$ number of PPO updates.
    \item \textbf{Replay-based}: $T$-step rollout length $\times$ number of parallel agents $\times$ (number of \texttt{DR} update-cycles $+$ number of \texttt{Replay} update-cycles $+$ number of \texttt{Mutation} update-cycles).
    \item \textbf{PAIRED}: $T$-step rollout length $\times$ number of parallel agents $\times$ number of PPO updates $\times$ number of students.
\end{itemize}

Note that our PAIRED implementation uses 2 students: the protagonist and antagonist; we exclude the number of interactions with the level editor environment in our measurements. 

\begin{figure}[h]
    \label{fig:minimax_levels}
    \centering
    \includegraphics[width=\linewidth]{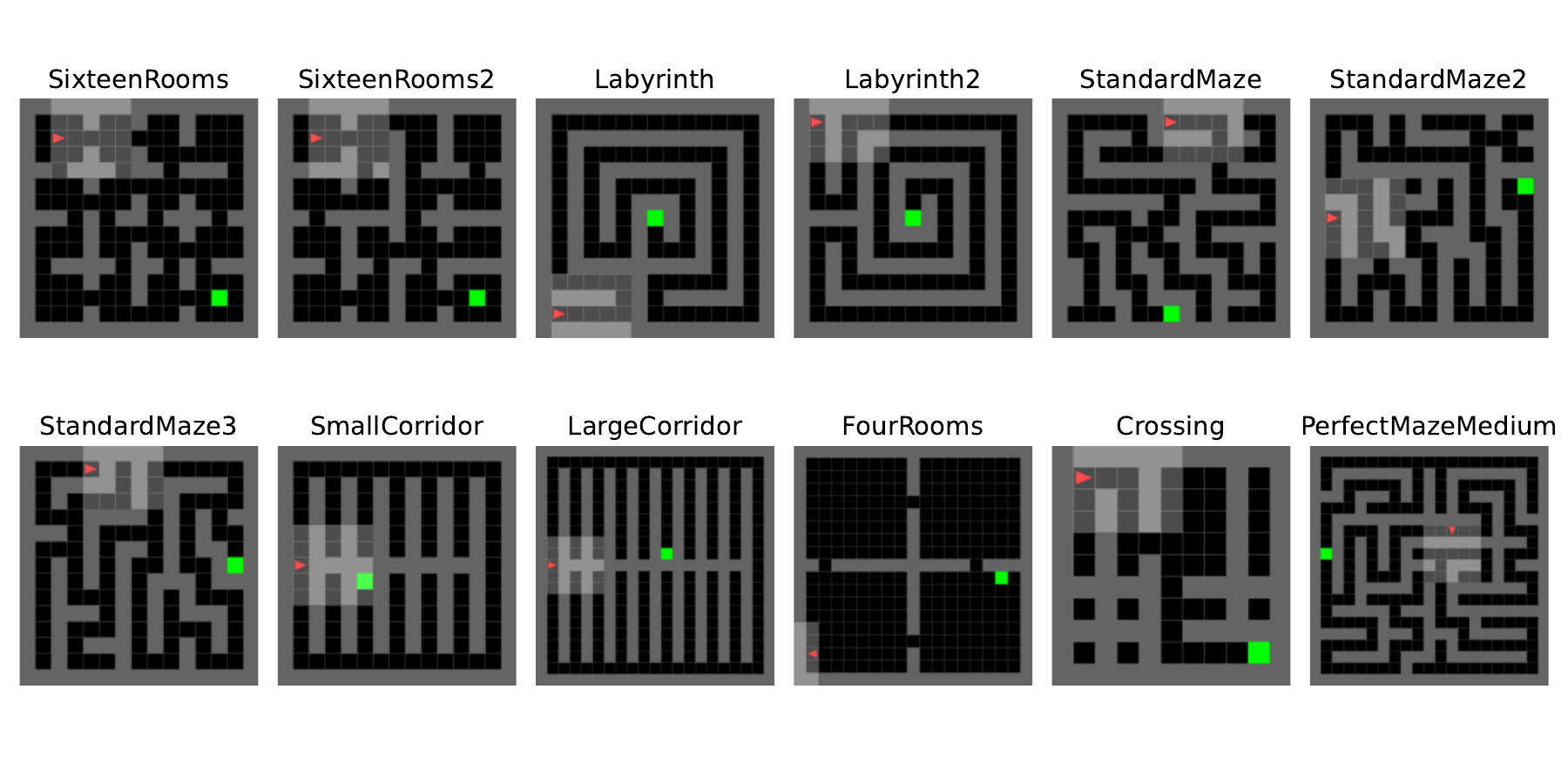}
    \caption{Visualization of an example batch of holdout levels used for evaluation, generated using \minimax \citep{jiang2023Minimax}. Such levels were used to evaluate the performance of \minimax and DCD in \citep{jiang2023Minimax}.}
    \label{fig:holdouts}
\end{figure}

\subsection{Performance}

We report the performance of each algorithm over exactly 245760000 environment interactions.\footnote{If all environment steps are used for training, this corresponds to the commonly-used 30k PPO updates of rollout length 256 with 32 parallel environments.} Each method uses an identical actor-critic network architecture and identical hyperparameters in common. We evaluate each method on a set of holdout levels, taken from prior work~\citep{jiang2021Replayguided,holder2022Evolving}, shown in Figure~\ref{fig:holdouts}. Hyperparameters used for training are listed in Table~\ref{table:ued:hyperparams}. We selected hyperparameters similar to those reported in Minimax \citep{jiang2023Minimax}.

\begin{figure}[h]
    \centering
    \includegraphics[width=\linewidth]{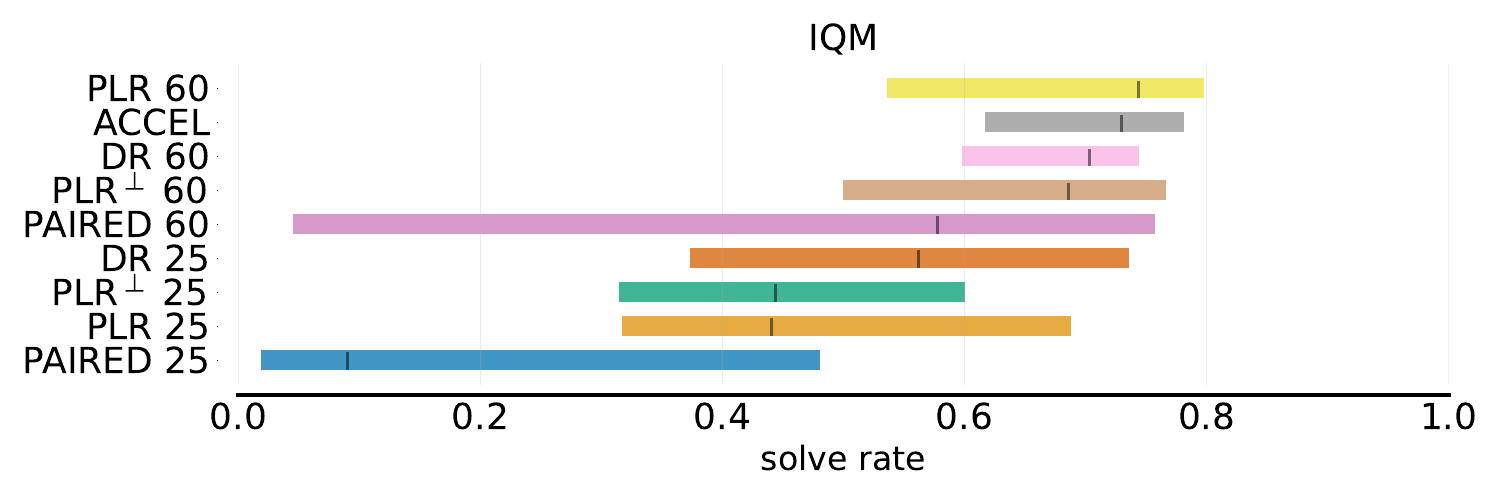}
    \caption{IQM of mean solve rate across over 100 trials of \minimax evaluation levels, measured over 10 random seeds. Error bars correspond to min-max performance over the seeds. The number after the method name indicates the maximum number of walls in the base DR distribution, either 25 or 60, and in the case of PAIRED indicates the number of editor environment steps taken by the adversary.}
    \label{fig:iqm_minimax}
\end{figure}

Surprisingly, our results, illustrated in Figure~\ref{fig:iqm_minimax}, demonstrate that, contrary to previous results \citep{dennis2020Emergent, jiang2020Prioritized, jiang2021Replayguided, holder2022Evolving, jiang2023Minimax}, DR is competitive with state-of-the-art UED methods. DR is also shown to significantly outperform any other UED method in restrict to a 25-wall budget. 

\subsection{Runtime Speed}

We report the wall clock time of each algorithm over exactly 245760000 environment interactions. All methods were run in identical environments using a single Nvidia A40. Note, these runs were measured with periodic evaluations enabled in addition to logging metrics to Weights and Biases every 2048000 environment steps: this includes rendering of levels and rollout animations on evaluation levels. Wallclock times are reported in Table~\ref{tab:wallclock_time}.

\section{Comparison to other libraries}

We validate our implementations by comparing each method's performance against prior implementations from other UED libraries, specifically DCD \citep{jiang2021Replayguided} and \minimax \citep{jiang2023Minimax}. For a fair comparison, we evaluate on the full holdout suite generated by \minimax. See Figure~\ref{fig:holdouts} for an example batch of procedurally generated evaluation levels.

\begin{table}[htbp]
\centering
\caption{Mean solve rate of each algorithm on the set of evaluation levels used by \citet{jiang2023Minimax}; measured over 10 random seeds. Error corresponds to standard deviation. dcd is the codebase released by \citet{jiang2021Replayguided} while \minimax is another Jax-based UED library~\citep{jiang2023Minimax}. *Previous implementations of PAIRED measure the number of environment steps only with respect to the protagonist, in effect treating antagonist environment steps as free. As such, results were collected using twice the number of environment steps as our previous analysis. **Previous implementations measure ACCEL performance on a fixed number of PPO updates where replay and mutation update cycles are bundled into a single update cycle. Consequently, the number of environment steps exceeds that of DR/PLR. } 
\label{tab:wallclock_time}
\begin{tabular}{l|ccccc}
\hline
& \textbf{DR} & \textbf{PAIRED}* & \textbf{PLR} & $\textbf{PLR}^\bot$ & \textbf{ACCEL}** \\
\hline
dcd (reported) & $0.62 \pm 0.05$ & $0.52 \pm 0.13$ & - & $0.71 \pm 0.04$ & $0.75 \pm 0.03$ \\
\minimax (reported) & $0.55 \pm 0.05$ & $0.63 \pm 0.04$ & - & $0.70 \pm 0.03$ & $0.73 \pm 0.05$ \\
\minimax + s5 policy (reported) & $0.58 \pm 0.05$ & $0.58 \pm 0.06$ & - & $0.66 \pm 0.04$ & $0.72 \pm 0.06$ \\
\hline
JaxUED & $0.69 \pm 0.05$ & $0.61 \pm 0.16$ & $0.72 \pm 0.08$ & $0.66 \pm 0.09$ & $0.72 \pm 0.05$ \\
JaxUED (25 wall limit) & $0.54 \pm 0.12$ & $0.17 \pm 0.16$ & $0.47 \pm 0.11$ & $0.46 \pm 0.09$ & - \\
\hline
\end{tabular}
\end{table}

\section{Conclusion}

This work introduces \jued, a minimal dependency, Jax-based UED library for researchers. We show that our implementations are able to reproduce prior results, but that DR is a surprisingly effective baseline contrary to findings from previous works. We hope that \jued will be useful in advancing the field of UED, with an overall aim of obtaining more general and robust RL agents.

\section{Acknowledgments and Disclosure of Funding}

Special thanks to Minqi Jiang, and authors of DCD and Minimax, whose implementations of UED methods heavily inspired this work. Thanks to Robert Lange, author of Gymnax, whose environment interface inspired the design of our \texttt{UnderspecifiedEnv} class. Thanks to Chris Lu whose hardware-accelerated implementations of RL greatly influenced this work. This work was funded by EPSRC Centre for Doctoral Training in Autonomous Intelligent Machines and Systems.

\bibliographystyle{abbrvnat}
\bibliography{bib}

\newpage
\appendix
\section{Hyperparams} 

\begin{table}[h!]
    \caption{Hyperparameters used for our experiments.}
    \label{table:ued:hyperparams}
    \begin{center}
    \scalebox{1.0}{
        \begin{tabular}{lr}
        \toprule
        \textbf{Parameter}              & Value \\
        \midrule
        \textbf{PPO}                    &           \\
        Number of Env Steps             & 245760000 \\
        $\gamma$                        & 0.995     \\
        $\lambda_{\text{GAE}}$          & 0.98      \\
        PPO number of steps             & 256       \\
        PPO epochs                      & 5         \\
        PPO minibatches per epoch       & 1         \\
        PPO clip range                  & 0.2       \\
        PPO \# parallel environments    & 32        \\
        Adam learning rate              & 1e-4      \\
        Anneal LR                       & yes       \\
        Adam $\epsilon$                 & 1e-5      \\
        PPO max gradient norm           & 0.5       \\
        PPO value clipping              & yes       \\
        return normalization            & no        \\
        value loss coefficient          & 0.5       \\
        entropy coefficient             & 1e-3      \\
        \textbf{Network}                  &           \\
        Number of Convolutional Filters & 16        \\
        Hidden Dimension                & 32        \\
        \textbf{PLR}                      &           \\
        Replay rate, $p$                & 0.5       \\
        Buffer size, $K$                & 4000      \\
        Scoring function                & MaxMC     \\
        Prioritisation                  & Rank      \\
        Temperature, $\beta$            & 0.3       \\
        Staleness coefficient           & 0.3       \\
        \textbf{ACCEL}                    &           \\
        Replay rate, $p$                & 0.8       \\
        Number of Edits                 & 20        \\
        \textbf{PAIRED}                   &           \\
        Adversary $\gamma$                    & 0.995     \\
        Adversary $\lambda_{\text{GAE}}$      & 0.98      \\
        Adversary epochs                      & 5         \\
        Adversary minibatches per epoch       & 1         \\
        Adversary clip range                  & 0.2       \\
        Adversary Adam learning rate          & 1e-4      \\
        Adversary Anneal LR                   & yes       \\
        Adversary Adam $\epsilon$             & 1e-5      \\
        Adversary max gradient norm           & 0.5       \\
        Adversary value clipping              & yes       \\
        Adversary return normalization        & no        \\
        Adversary value loss coefficient      & 0.5       \\
        Adversary entropy coefficient         & 5e-2      \\
        Adversary Number of Convolutional Filters & 128        \\
        Adversary Hidden Dimension                & 32        \\
        \bottomrule 
        \end{tabular}}
    \end{center}
\end{table}

\end{document}